# Circle detection using Electromagnetism-Like Optimization

*Erik Cuevas, *Diego Oliva, *Daniel Zaldivar[1], *Marco Pérez-Cisneros and +Humberto Sossa

*Departamento de Ciencias Computacionales
Universidad de Guadalajara, CUCEI
Av. Revolución 1500, Guadalajara, Jal, México
{erik.cuevas,diego.oliva,[1]daniel.zaldivar, marco.perez}@cucei.udg.mx
+Centro de Investigación en Computación, IPN
Av. Juan de Dios Bátiz s/n, Mexico DF.
hsossa@cic.ipn.mx

**Abstract**

This paper describes a circle detection method based on Electromagnetism-Like Optimization (EMO). Circle detection has received considerable attention over the last years thanks to its relevance for many computer vision tasks. EMO is a heuristic method for solving complex optimization problems inspired in electromagnetism principles. This algorithm searches a solution based in the attraction and repulsion among prototype candidates. In this paper the detection process is considered to be similar to an optimization problem, the algorithm uses the combination of three edge points ($x$, $y$, $r$) as parameters to determine circles candidates in the scene. An objective function determines if such circle candidates are actually present in the image. The EMO algorithm is used to find the circle candidate that is better related with the real circle present in the image according to the objective function. The final algorithm is a fast circle detector that locates circles with sub-pixel accuracy even considering complicated conditions and noisy images.

## 1. Introduction

The problem of detecting circular features holds paramount importance for image analysis, in particular for industrial applications such as automatic inspection of manufactured products and components, aided vectorization of drawings, target detection, etc. [1]. Solving object location challenges is normally approached from two types of techniques: deterministic techniques which include application of Hough transform based methods [2], geometric hashing and template or model matching techniques [3, 4]. In the other hand, stochastic techniques include random sample consensus techniques [5], simulated annealing [15] and Genetic Algorithms (GA) [6].

Template and model matching techniques were the first approaches applied to shape detection. Plenty of methods have been developed to solve the shape detection problem [7]. Shape coding techniques and combination of shape properties were used to represent such objects. The main drawback of these techniques is related to the contour extraction step from real images. Additionally, it is difficult for models to deal with pose invariance except for very simple objects.

Commonly, the circle detection in digital images is realized by means of Circular Hough Transform [8]. A typical Hough-based approach employs an edge detector and uses edge information to infer locations and radius values. Peak detection is then performed by averaging, filtering and histogramming the transform space. However, such approach requires a large storage space given the 3-D cells needed to cover the parameters ($x$, $y$, $r$), the computational complexity and the low processing speed. The accuracy of the extracted parameters of the detected circle is poor, particularly in presence of noise [13]. Specifically, for a digital image of significant width and height and densely populated edge pixels, the required processing time for Circular Hough Transform makes it prohibitive to be deployed in real time applications. In order to overcome such problem, some other researchers have proposed new approaches based in the Hough transform. We can find in the literature, for example, the probabilistic Hough transform [9], the randomized Hough transform (RHT) [10], the fuzzy Hough transform [11]. We can even find alternative transforms as proposed by Becker [12]. Although those new approaches show better processing speeds (in comparison with the original Hough Transform), they continue being high sensitive in noise presence.

---
[1] Corresponding author, Tel +52 33 1378 5900, ext. 7715, E-mail: daniel.zaldivar@cucei.udg.mx

Another alternative for the shape recognition in computer vision has been the stochastic search methods such as the Genetic Algorithms (GA). In particular, GA has recently been used for important shape detection task e.g. Roth and Levine proposed use of GA for primitive extraction of images [16]. Lutton et al., carried out a further improvement of the aforementioned method [17]. Yao *et al.,* came up with a multi-population GA to detect ellipses [18]. In [20], GA was used for template matching when the pattern has been the subject of an unknown affine transformation. Ayala–Ramirez *et al.,* presented a GA based circle detector [19] which is capable of detecting multiple circles on real images but fails frequently to detect imperfect circles. Another example is presented in [21] where is employed soft computing techniques applied to shape classification. In the case of ellipsoidal detection, Rosin proposes in [22] an ellipse fitting algorithm that uses five points. In [23], Zhang and Rosin extends the later algorithm to fit data on super-ellipses. La mayoría de estos enfoques permite la detección de círculos en presencia de ruido y con velocidades de procesamiento aceptables. Sin embargo, fallan en la detección de círculos en condiciones complejas tales como oclusión y superposición.

EMO algorithm [10] is a stochastic evolutionary computation technique based on the electromagnetism theory in physics. The convergence property of EMO algorithm has already been proved in [11,45]. EMO algorithm is an intelligent technique different from GA and Simulated Annealing (SA). The way in which the particles of algorithm move is correspondent with particle swarm optimization (PSO) [12] and Ant Colony Optimization (ACO) [13]. The first step of EM-like algorithm is to produce one group random solution from feasible domain, and regard each solution as a charged particle. The charge of each particle is determined by the fitness function, and then moves the particle with attraction or repulsion among population. The attraction–repulsion mechanism of EMO algorithm corresponds to the reproduction, crossover and mutation in GA [14].

EMO algorithm calculates the resultant force in the population to determine the moving direction of the current particle by Coulomb's law and superposition principle. The resultant force is decided by the charges and distance among each particle. In this mechanism, the higher charge will produce larger attraction or repulsion. The resultant force is small when the distance between the particles is farther. In latter iteration, the movements of particles will be slow, as those of SA [14]. On the other hand, EMO algorithm can improve the current optimum solution with local search and advances the feasibility for global search.

In general, the EMO algorithm can be considered as a fast and robust algorithm representing a real alternative to solve complex, nonlinear, non-differentiable and non-convex optimization problems. The principal advantages of the EM algorithm are: it has no gradient operation, it can be used in decimal system directly, it needs a few particles to converge and the proof of convergence has been already verified [45].

This paper presents a circle detector method based in the Electromagnetism-Like algorithm. Here, the detection process is considered to be similar to an optimization problem. The algorithm employs a three edge point circle representation that lets the system to reduce the search space by eliminating unfeasible circle locations in the image. This approach results in a sub-pixel circle detector that can effectively identify circles in real images even considering complicated conditions. El artículo muestra evidencia experimental relevante, en la detección de círculos ante condiciones de oclusión, imperfección, existencia parcial (arcos) e inclusive dibujados a mano.

This paper is organized as follows: Section 2 provides a brief outline of the EM theory. In Section 3, we formulate our approach and we also present the main characteristics of the EM algorithm used to detect circles in images. Section 4 shows the results of applying our method to the recognition of circles in different image conditions. Our conclusions are presented in Section 5; we also discuss here some future work to be done.

## 2. Electromagnetism - Like Algorithm

In this paper, EM-like algorithm is utilized to detect circles presented in an image. The EM algorithm is a simple and direct search algorithm based on population that allows optimizing global multi-modal functions. In comparison with genetic algorithm, it is not employs the crossover and mutation operators for exploration of the feasible regions. The algorithm is based on physical principles.

The EM algorithm allows solving a special class of optimization problems with bounded variables in the form of:

$$\min f(x) \\ x \in [l, u] \qquad (1)$$

where $[l, u] = \{x \in \Re^n \mid l_d \leq x_d \leq u_d, d = 1, 2...n\}$ and $n$ the dimension of the variable $x$.

EM-like algorithm has four main phases [10], i.e., initialization, local search, calculation and movement, respectively. They are described as following:

**Initialize,** $m$ particles are taken considering the upper ($u$) and lower limit ($l$).
**Local search,** it is used to gather the local information for a point $x^p$, where $p \in (1,\ldots,m)$.
**Calculation of the total force vector,** the charges and the forces are calculated for every particle.
**Movement,** each particle is moved according to the resultant force.

*2.1 Initialize*

At first, one group **G** of $m$ solutions $n$-dimensional is randomly produced as initial state. Each solution is regarded as a charged particle and all particles are assumed to be uniformly distributed between the upper ($u$) and lower ($l$) bounds. The optimum particle in the population will then be found by the fitness function which depends of the optimization problem.

*2.2 Local Search*

Local search should be able to find better solution in theory. However local search may be unnecessary for some problems. The algorithms can be classified as EM-like without local search, EM-like with local search applied to all particles and EM-like with local search applied to the current better particle only.

Considering a determined number of iterations *ITER* and a feasible neighborhood search $\delta$, the procedure iterates as follows: For a given coordinate $d$, the point $x^p$ is assigned to a temporary point $y$ to store the initial information. Next, a random number is selected and combined with $\delta$ as a step length, thus the point $y$ is moved along that direction where the sign is also determined randomly. If the point $y$ observes a better point within *ITER* iterations, the point $x^p$ is replaced by $y$ and the neighborhood search for point $p$ ends. Finally the *current best* point is updated. The algorithm in pseudo-code is shown in Fig. 1.

In general, local search to all particles can reduce the risk that falls in local solution but it is relatively time consuming. Local search only on the current better particle is a better choice to maintain the computational efficiency and precision. In addition, the step length for local search is an important factor that depends on the bounds of each dimension. Generally speaking, the step length determines the performance of local search.

*2.3 Calculation of The Total Force Vector*

The calculation of the total force vector is based in the *superposition principle* (Figure 2) of electromagnetism theory which states that the force exerted on a point via other points is inversely proportional to the distance between the points and directly proportional to the product of their charges

[46]. The particle moves according to Coulomb's force produced among the particles, as we assign a charge-like value to each particle. The charge of each particle is determined by its fitness function value, which can be evaluated as

$$q^p = \exp\left(-n \frac{f(x^p) - f(x^{best})}{\sum_{h=1}^{m}(f(x^h) - f(x^{best}))}\right), \forall p \tag{2}$$

where $n$ denotes the dimension and $m$ represents the population size in EM-like algorithm. Higher dimension usually requires larger population. In Eq. (2), the particle with the best fitness function $x^{best}$ value is called "best particle", and will have the highest charges. A particle will have stronger attraction, as it appears near the optimum particle. The particle attracts other particles with better fitness function values, and repels other particles with worse fitness function values.

```
1:  counter ← 1                          11: y_d ← y_d − λ_2·δ
2:  for p = 1 to m do                    12: end if
3:  for d = 1 to n do                    13: if f(y) < f(x^p)) then
4:  λ_1 ← U(0, 1)                        14: x^p ← y
5:  while counter <ITER do               15: end if
6:  y ← x^p                              16: counter ← counter + 1
7:  λ_2 ← U(0, 1)                        17: end while
8:  if λ_1 > 0.5 then                    18: end for
9:  y_d ← y_d + λ_2·δ                    19: end for
10: else                                 20: x^{best} ← arg min{f(x^p), ∀p}
```

**Fig. 1.** Pseudo-code of the local search procedure.

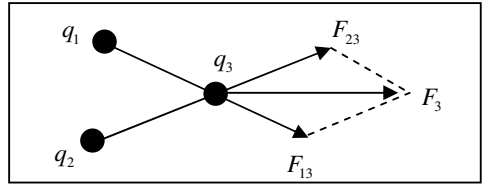

**Fig. 2.** The superposicion principle.

The resultant force among particles determines the effect for optimization process. The resultant force of each particle can be evaluated by Coulomb's law and superposition principle as

$$F^p = \sum_{h \neq p}^{m} \begin{cases} (x^h - x^p)\dfrac{q^p q^h}{\|x^h - x^p\|^2} & \text{if } f(x^h) < f(x^p) \\ (x^p - x^h)\dfrac{q^p q^h}{\|x^h - x^p\|^2} & \text{if } f(x^h) \geq f(x^p) \end{cases}, \forall p \tag{3}$$

Where $f(x^h) < f(x^p)$ represents attraction and $f(x^h) \geq f(x^p)$ represents repulsion. The resultant force of each particle is proportional to the product of the charges and is inversely proportion to the distance between the particles. In order to maintain the feasibility, Eq. (3) should be normalized as

$$F^p = \frac{F^p}{\|F^p\|}, \quad \forall p. \tag{4}$$

*2.4 Movement.*

Each particle moves according to the resultant force which can be given as

$$x^p = \begin{cases} x^p + \lambda \cdot F^p \cdot (u_d - x_d^p) & if \quad F^p > 0 \\ x^p + \lambda \cdot F^p \cdot (x_d^p - l_d) & if \quad F^p \leq 0 \end{cases}, \forall p \neq best \tag{5}$$

In Eq. (5), $\lambda$ is a random step length, which is uniformly distributed between zero and one. Where $u_d$ and $l_d$ represent the upper and lower bounds for the *d*-dimension, respectively. The particle moves toward the upper bound by a random step length as the resultant force is positive, or moves toward the lower bound as the resultant force is negative. The best particle does not move, because it is a particle with absolute attraction which attracts all other particles in the population.

**3. Circle detection using Electromagnetism-Like**

Circles are represented in this work by means of parameters of the well-known second degree equation (see Equation 6), that passes through three points [13] in the edge space of the image. Images are preprocessed by an edge detection step which uses a single-pixel edge detection method for object's contour. This task is accomplished by the classical Canny algorithm which stores the locations for each edge point. Thus, such points are the only potential candidates to define circles by considering triplets. All the edge points in the image are then stored within a vector array $E = \{e_1, e_2, \ldots, e_{Np}\}$ with *Np* as the total number of edge pixels contained in the image. The algorithm stores the $(x_v, y_v)$ coordinates for each edge pixel $e_v$ in the edge vector.

In order to construct each of the circle candidates (or particles within the EM framework), the indexes $v_1$, $v_2$ and $v_3$ of three edge points must be combined, assuming the circle's contour goes through points $e_{v_1}$; $e_{v_2}$; $e_{v_3}$. A number of candidate solutions are generated randomly for the initial pool. The solutions will thus evolve through the application of the EM algorithm as the evolution takes place over the pool until a minimum is reached and the best individual is considered as the solution for the circle detection problem.

Applying classic methods based on Hough Transform for circle detection would normally require huge amounts of memory and consume large computation time. In order to reach a sub-pixel resolution –an equal feature of the method presented in this paper– they also consider three edge points to cast a vote for the corresponding point within the parameter space. Such methods also require an evidence-collecting step how is also implemented by the method in this paper. However as the overall evolution process evolves, the objective function improves at each generation by discriminating non-plausible circles and locating others by avoiding a visit to other image points. The discussion that follows, clearly explains the required steps to formulate the circle detection task just as an EM optimization problem.

*3.1. Individual representation*

Each particle *C* of the pool uses three edge points as elements. In this representation, the edge points are stored according to one index that is relative to their position within the edge array *E* of the image. In turn, the procedure will encode a particle as the circle that passes through three points $e_i$, $e_j$ and $e_k$ ($C = \{e_i, e_j, e_k\}$). Each circle *C* is represented by three parameters $x_0$, $y_0$ and *r*, being $(x_0, y_0)$ the coordinates of the center of the circle and $r$ its radius. The equation of the circle passing through the three edge points can thus be computed as follows:

$$(x - x_0)^2 + (y - y_0)^2 = r^2 \tag{6}$$

considering

$$\mathbf{A} = \begin{bmatrix} x_j^2 + y_j^2 - (x_i^2 + y_i^2) & 2 \cdot (y_j - y_i) \\ x_k^2 + y_k^2 - (x_i^2 + y_i^2) & 2 \cdot (y_k - y_i) \end{bmatrix} \mathbf{B} = \begin{bmatrix} 2 \cdot (x_j - x_i) & x_j^2 + y_j^2 - (x_i^2 + y_i^2) \\ 2 \cdot (x_k - x_i) & x_k^2 + y_k^2 - (x_i^2 + y_i^2) \end{bmatrix}, \tag{7}$$

$$x_0 = \frac{\det(\mathbf{A})}{4((x_j - x_i)(y_k - y_i) - (x_k - x_i)(y_j - y_i))}, \quad y_0 = \frac{\det(\mathbf{B})}{4((x_j - x_i)(y_k - y_i) - (x_k - x_i)(y_j - y_i))}, \tag{8}$$

and

$$r = \sqrt{(x_0 - x_b)^2 + (y_0 - y_b)^2}, \tag{9}$$

being det(.) the determinant and $b \in \{i, j, k\}$. Figure 2 illustrates the parameters defined by Equations 7 to 9.

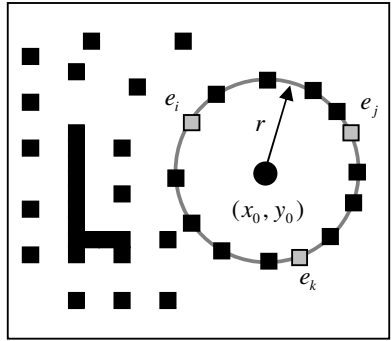

**Fig. 2.** Circle candidate (individual) built from the combination of points $p_i$, $p_j$ and $p_k$.

Thus, it is possible to represent the shape parameters (for the circle, $[x_0, y_0, r]$) as a transformation $T$ of the edge vector indexes $i, j$ and $k$.

$$[x_0, y_0, r] = T(i, j, k) \tag{10}$$

with $T$ being the transformation calculated after the previous computations for $x_0$, $y_0$, and $r$.

By exploring each index as an individual parameter, it is possible to sweep the continuous space looking for the shape parameters using the EM optimization algorithm. This approach reduces the search space by eliminating unfeasible solutions.

*3.2 Objective function*

A circumference may be calculated as a virtual shape in order to measure the matching factor between $C$ and the presented circle in the image. It must be also validated, i.e. if it really exists in the edge image. The test for these points is $S = \{s_1, s_2, \ldots, s_{Ns}\}$, with $Ns$ representing the number of test points over which the existence of an edge point will be verified.

The test $S$ is generated by the midpoint circle algorithm [37]. The midpoint circle algorithm (MCA) determines the required points for drawing a circle, considering the radius $r$ and the center point $(x_0, y_0)$. The algorithm employs the circle equation $x^2 + y^2 = r^2$, with only the first octant. It draws a curve starting at point (*r*, 0) and proceeds upwards-left by using integer additions and subtractions. See full details in [38].

The MCA aims to calculate the points *Ns* which are required to represent the circle considering coordinates $S = \{s_1, s_2, \ldots, s_{Ns}\}$. Although the algorithm is considered the quickest providing a sub-pixel precision, it is important to assure that points lying outside the image plane must not be considered as they must be included in *Ns*, thus protecting the MCA operation.

The matching function or objective function *J(C)* represents the matching (or error) resulting from pixels *S* for the circle candidate and the pixels that really exist in the edge image, yielding:

$$J(C) = 1 - \frac{\sum_{v=1}^{Ns} E(x_v, y_v)}{Ns} \qquad (11)$$

with $E(x_v, y_v)$ accumulating the number of expected edge points (the points in *S*) that are actually present in the edge image. *Ns* is the number of pixels within the perimeter of the circle that correspond to *C*, currently under testing.

Therefore the algorithm aims to minimize *J(C)*, given that a smaller value implies a better response (matching) of the "circularity" operator. The optimization process can thus be stopped after the maximum number of epochs is reached and the individuals are clearly defined satisfying the threshold. The stopping criterion depends on the a priori knowledge about the application context.

*3.3. EM Implementation*

The implementation of the proposed algorithm can be summarized into the following steps:

**Step 1:** Se aplica el filtro de canny a la imagen para obtener sus bordes. Los puntos que constituyen los bordes de la imagen son almacenados en el vector $E = \{e_1, e_2, \ldots, e_{Np}\}$, además de coloca *iteration*=0.

**Step 2:** Considerando la fase de inicialización de EM, se generan *m* partículas con tres elementos cada una $e_i, e_j$ y $e_k$, donde $e_i, e_j$ y $e_k \in E$. Dichas partículas conforman el conjunto inicial de partículas. Cada elemento de *C* contiene uno de los índices del vector *E*. Son eliminadas las partículas que presenten radios muy grandes o muy pequeños. It is also evaluated the objective value of $J(C^p)$ for all individuals and determined the best particle $C^{best}$, where $C^{best} \leftarrow \arg\min\{J(C^p), \forall p\}$.

**Step 3:** Considering a determined number of iterations *ITER* and a feasible neighborhood search $\delta$. For a given coordinate *i*, *j* or *k*, the particle $C^p$ is assigned to a temporary point *y* to store the initial information. Next, a random number is selected and combined with $\delta$ as a step length, thus the point *y* is moved along that direction where the sign is also determined randomly. If the point *y* observes a better point within *ITER* iterations, the particle $C^p$ is replaced by *y* and the neighborhood search for particle *p* ends. Finally the *current best* particle $C^{best}$ is updated.

**Step 4:** Se calcula la carga ejercida entre las partículas, utilizando la ecuación (2), después aplicando las cargas calculadas en la ecuación (3), se obtiene el vector de que contiene la fuerza generada entre las partículas, tomando en cuenta que la partícula que tiene un mejor valor en la función de costo tendrá también una mayor carga y por lo tanto una mayor fuerza de atracción.

**Step 5:** Se desplazan las partículas, de acuerdo a la magnitud de su fuerza. La nueva posición de la partícula es calculada a partir de la ecuación (4), $C^{best}$ no es desplazada ya que al tener la mayor fuerza atrae a las demás partículas.

**Step 6:** Se incrementa *iteration* y se prueba la condición de paro del algoritmo que puede ser que transcurriera un máximo numero de iteraciones (*iteration = MAXITER*) o que *J(C)* es mas pequeño que un umbral definido. Si se cumple alguna de las condiciones se pasa al paso7, si no se va al paso 3.

**Step 7** Se toma la mejor partícula $C^{best}$ de la última iteración.

**Step 8:** Del mapa de bordes original se eliminan los puntos encontrados por el algoritmo correspondientes a la partícula $C^{best}$, y en caso de buscar mas de un círculo se regresa al paso 2, y se relaza el mismo procedimiento para cada uno de los círculos que se pretende encontrar.

**Step 9:** Por ultimo se toma la mejor partícula $C^{best}_{Nc}$ de cada círculo buscado, donde *Nc* es numero total de círculos buscados y se dibuja en la imagen original para mostrar los círculos encontrados por el algoritmo.

## 4. Experimental results

In order to evaluate the performance of the circle detector proposed in this paper, several experimental tests have been developed as follows:

(4.1) Circle detection
(4.2) Shape discrimination
(4.3) Multiple circle detection
(4.4) Circular approximation
(4.5) Approximation from occluded circles, imperfect circles or arc detection

Para todos los experimentos se toma una población de partículas de *m=10,* un numero máximo de iteraciones para la búsqueda local *LSITER=2,* la longitud de paso para la búsqueda locas δ = 3, y como criterio de termino del algoritmo un máximo de iteraciones *MAXITER=20,* El espacio de búsqueda esta comprendido como la cantidad de píxeles que forman el borde de la imagen el vector queda comprendido entre los limites *u=1,l= Np*, para cada variable $e_i$, $e_j$ y $e_k$.

*4.1 Circle localization*

4.1.1. Synthetic images

The experimental setup includes the use synthetic images of 200x200 pixels. Each image has been generated drawing only an imperfect circle (ellipse), randomly located. Some of these images were contaminated adding noise to increase the complexity in the detection process. The parameters to be detected are the center of the circle position (*x, y*) and its radius (*r*). The algorithm was set to 20 iterations for each test image. In all the cases the algorithm was able to detect the parameters of the circle, even in presence of noise. The detection is robust to translation and scale conserving a reasonably low elapsed time (typically under 1ms). Figure 3 shows the results of the circle detection for two different synthetic images.

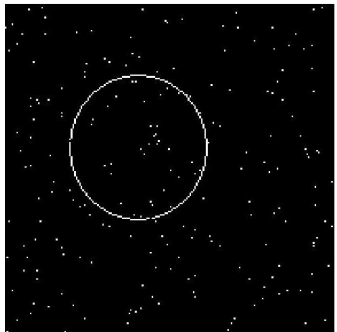
(a)

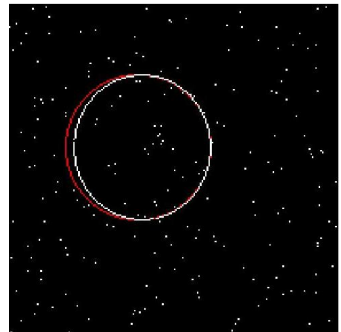
(b)

**Fig. 3.** Circle detection from synthetic images: (a) an original circle image with noise,
(b) its corresponding detected circle .

4.1.2. Natural images

This experiment tests the circle detection upon real-life images. Twenty five images of 640x480 pixels are used on the test. All are captured using digital camera with a 8 bits color format. Each natural scene includes a circle shape among other objects. All images are preprocessed using an edge detection algorithm and then feeding them into the EM-based detector. Figure 4 shows two particular cases from the 25 test images.

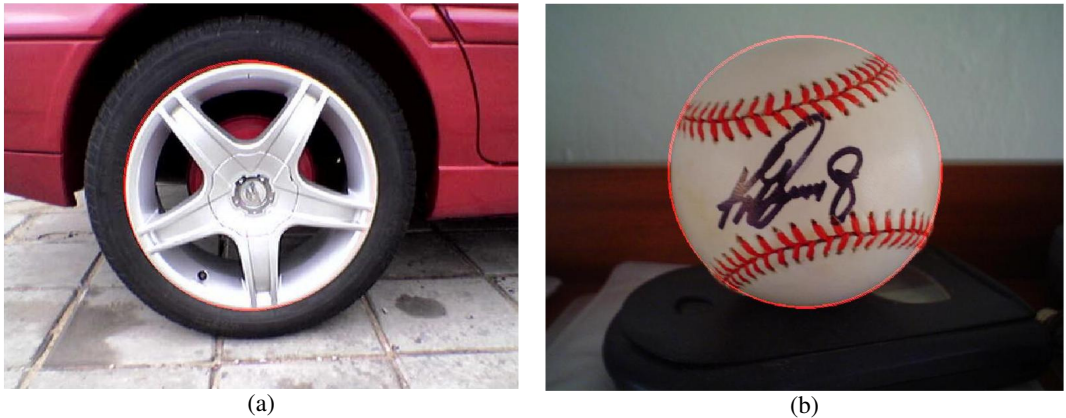

(a)           (b)

**Fig. 4.** Circle detection applied to a real-life images.(a) detected circle is shown near the ring periphery. (b) The detected circle is shown near the ball's periphery.

Real-life images rarely contain perfect circles, and therefore the detection algorithm approximates the circle that better adapts to the imperfect circle within the noisy image. Such circle corresponds to the smallest error obtained for the objective function *J(C)*. The results on detection have been statistically analyzed for comparison purposes. The times for detection in each image are 13.540807 seconds for Figure 4 (a), and 27.020633 seconds for Figure 4 (b). For instance, the detection algorithm was executed 20 times on the same image (Figure 4 (b)), resulting always the same parameters $x_0 = 214$, $y_0 = 322$, and *r* = 948. This indicates that the proposed EM algorithm is able to converge to the minimum solution obtained from the objective function *J(C)*. For this experiment 20 iterations were used.

*4.2. Shape discrimination tests*

This section discusses on the circle detection ability when any other shapes are present in the image. Five synthetic images of 540x300 pixels are considered for this experiment. Noise has been added to all images. A maximun of 20 iterations was used in the detection process; different shapes are also present in the images. Two examples are shown in the Figure 5.

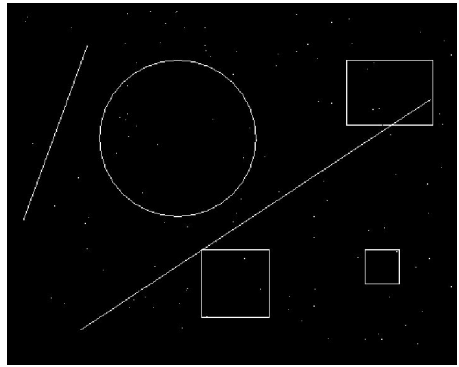
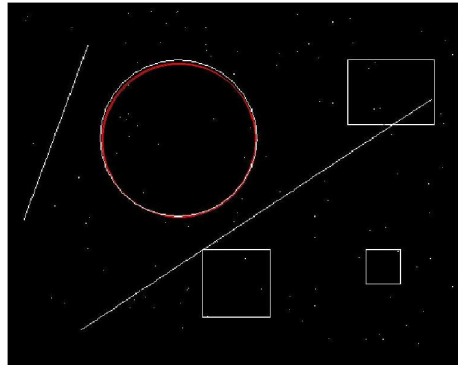

(a)                      (b)

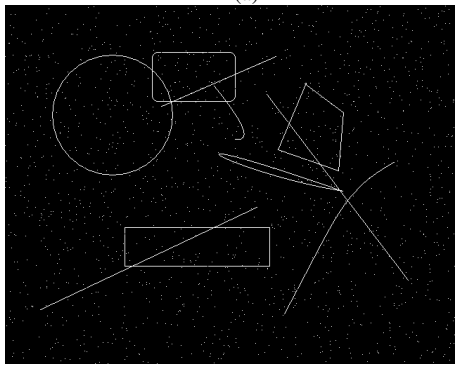
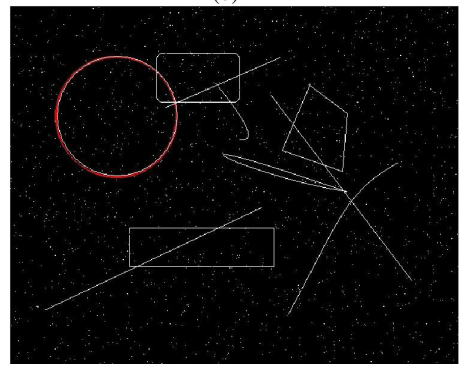

(c)                      (d)

**Fig. 5** Samples of synthetic images, (a) and (c) are the original images, (b) and (d) the detected circles.

The same experiment is repeated using real-life images, for instance the images shown in Figure 6. The circle detection from other shapes is completely feasible on natural real-life images.

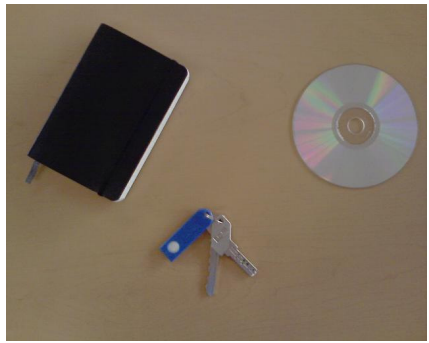
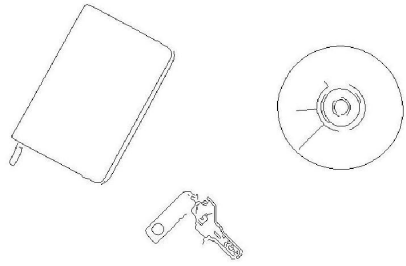

(a)                      (b)

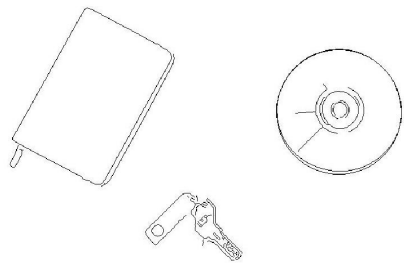
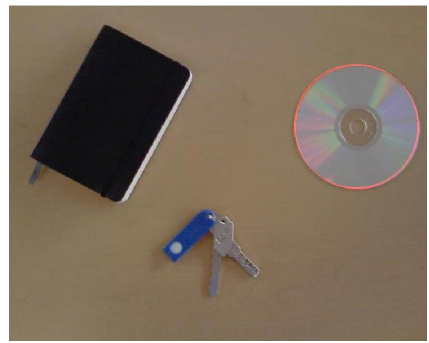

(c) (d)

**Fig. 6.** Different shapes embedded into a real-life image. (a) The test image (b) the corresponding edge map (c) circle detected and (d) the detected circle over the original image.

*4.3. Multiple circle detection*

The approach is also capable of detection several circles within a real-life image. It is necessary to set a maximal number of circular shapes to be found. Then the approach will work on the original edge image until the first circle is detected. The first circle represents the circle with the minimum objective function value $J(C)$. This shape is then masked (eliminated) on the edge image and the EM circle detector operates over the modified image. This procedure is repeated until the maximum number of detected shapes is reached. Finally, a validation of all detected circles is performed by analyzing continuity of the detected circumference segments as proposed in [41]. Such procedure is necessary because we could want to detect more circular shapes than the number of circles actually present in the image. The system can then provide a false response: "no circle detected", if none of the detected shapes satisfies circular completeness. The algorithm also seeks to identify any other circle-like shapes in the image selecting the best shapes up until the maximum number that it has been defined before for the algorithm.

Figure 7(a) shows the edge image obtained after application of the Canny algorithm as preprocessing. Figure 7(b) shows a real image including several detected circles which have been sketched on it. The same cases are for Figure 7(c) and 7(d).

The EM algorithm is an iterative procedure which allows identifying circular shapes considering an edge map containing the potential candidates. Each circle is rated according to the value of the objective function $J(C)$, which keeps tracking a number of circles that is defined at the beginning of the algorithm.

The iterative nature of the EM algorithm means that the input image to the current step in the algorithm is the optimized image from the previous step. The last image therefore does not longer include any fully detected circle because it has been already detected and it is not longer considered for future steps. The algorithm moves on focusing exclusively on other potential maps which may or may not represent another circle. A maximum of 20 iterations is normally considered as the limit for detecting a potential circle.

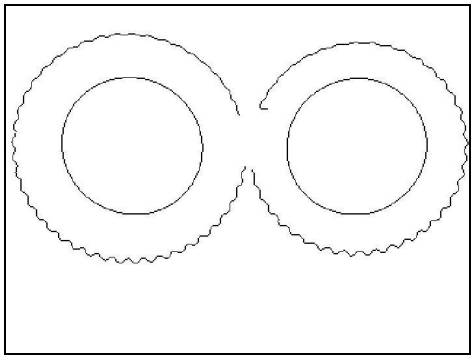 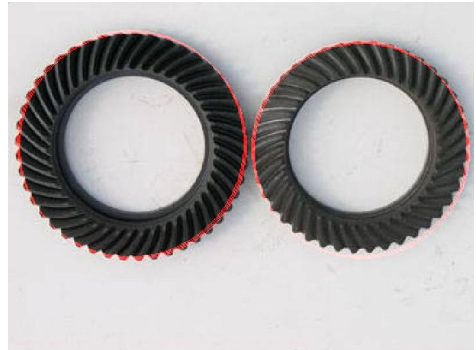

(a) (b)

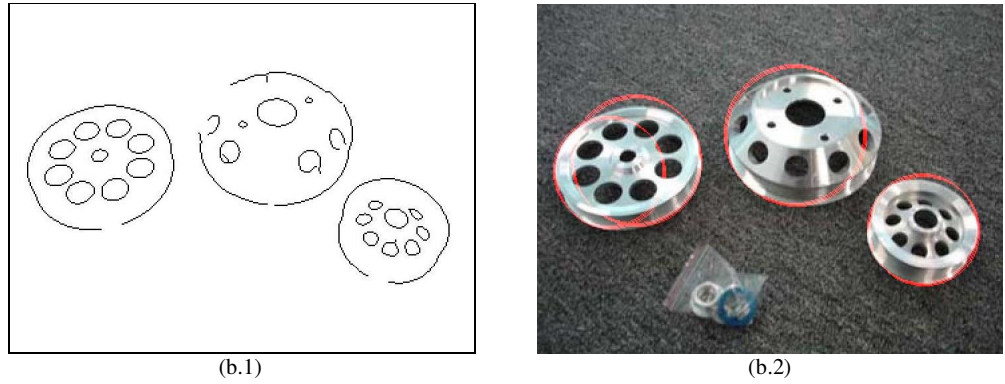

(b.1)                (b.2)

**Fig. 7.** Multiple circle detection on real images: (a) and (b) edge images obtained from applying the Canny algorithm, (c) and (d) Original images with the detected circles drawn over it.

*4.4 Circular approximation*

Since circle detection has been considered as an optimization problem, it is possible to approximate a given shape as the concatenation of circles. This can be achieved using the feature of the EM algorithm which may detect multiple circles just as it was explained in the previous sub-section. Thus, the EM algorithms may continually find circles which may approach a given shape according to the values obtained by the objective function $J(C)$.

Figure 8 shows some examples of circular approximation. En la figura 8(a) se tiene una forma constituida de un conjunto de círculos superpuestos, mientras que en la 8(b) se muestra la aproximación circular obtenida por el algoritmo propuesto. En la figura 8(c) se tiene una elipse, mientras que en la 8(d) la aproximación circular obtenida con 4 circulos.

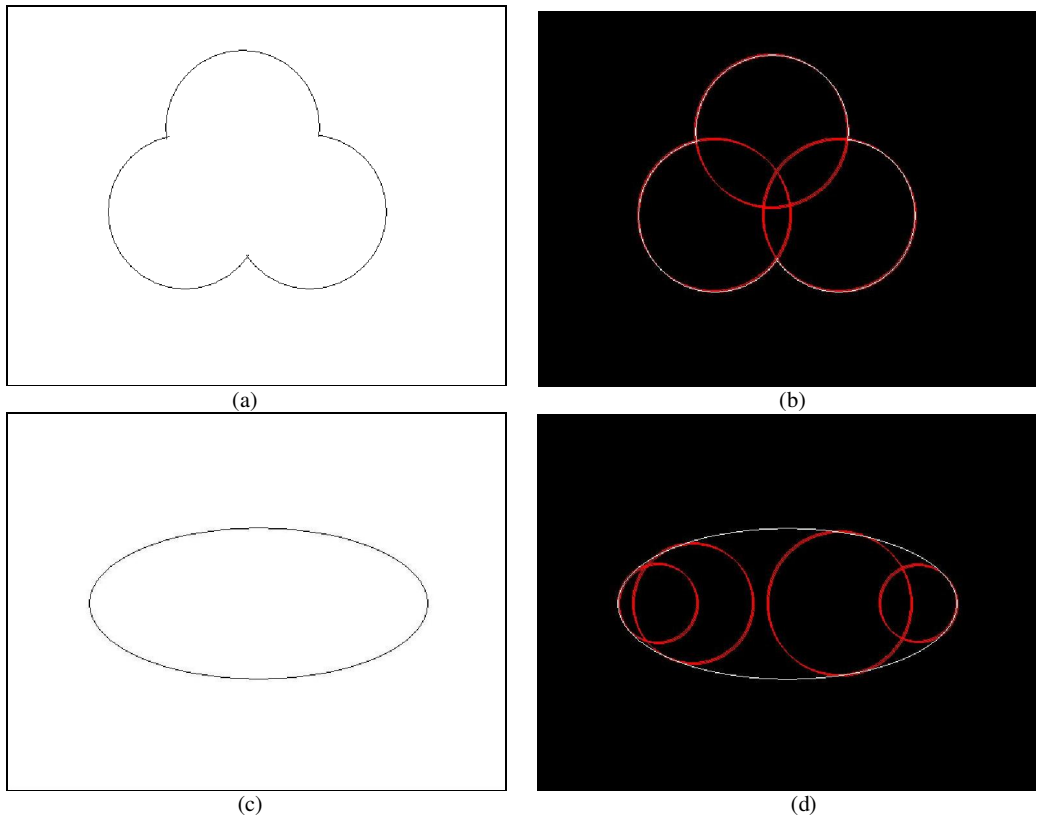

(a)                (b)

(c)                (d)

**Fig. 8.** Circular approximation: (a) The original image (b) its circular approximation considering 3 circles, (c) original image, and (d) its circular approximation considering 4 circles.

*4.4.2 Circle extraction from occluded or imperfect circles and arc detection.*

Circle detection may also be useful to approximate circular shapes from arc segments, occluded circular shapes or imperfect circles, all commonly found in typical computer vision problems. The EM algorithm may naturally find shapes that approach a circle arc according to the values in the objective function *J(C)*. Figure 8 shows some examples of this functionality.

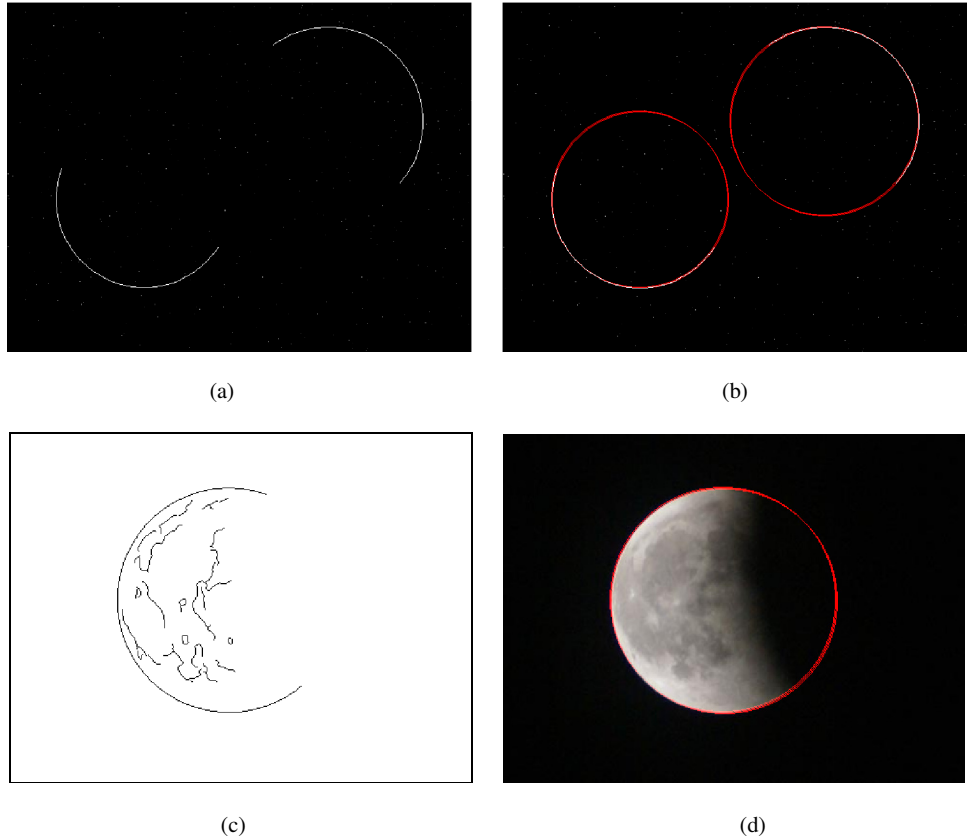

(a) (b)

(c) (d)

**Fig. 9.** Approximation of circles from occluded shapes, imperfect circles or arc detection: (a) original image with 2 arcs, (b) circle approximation for the first image, (c) Occluded natural image of the moon, (d) circle approximation for the moon example.

**5. Conclusions**

This paper has presented a circle detection method based on Electromagnetism - Like (EM) optimization. The approach is capable of detecting circles with sub-pixel accuracy in synthetic and natural images. The approach encodes circles based on three edge points by using the circle equation. The algorithm uses the combination of three edge points as parameters to determine circles candidates in the scene, reducing the search space and producing a fast circle detector which can be compared to other on the literature [2, 4, 12, 19]. The circle detector can reliably detect circular shapes even if significant occlusions, discontinuities or partial arcs are presented to the algorithm. The circle detector can reliably detect circular shapes even if significant occlusions, handwriting circles, discontinuities or partial arcs are presented to the algorithm.

Although the Hough Transform methods for circle detection also use three edge points to cast one vote for the potential circular point in the parameter space, they would require huge amounts of memory and longer computational time to obtain a sub-pixel resolution. Such methods also require an evidence-collecting step that is also implemented by the method in this paper but as the evolution process is performed and the objective function improves at each generation by discriminating non plausible circles. Thus, the circle will be located, without visiting several image points.

It is important to note that the use of the MCA in this paper exhibits more precision than the presented by Ayala-Ramirez et al. in [19]. Such procedure employs samples taken from circular patterns and the test $S$ is generated by the uniform sampling of the shape boundary. Each point $s_i$ is a 2D feature and its coordinates $(x_i, y_i)$ are computed as follows:

$$\begin{aligned} x_{s_i} &= x_0 + r \cdot \cos\left(\frac{2\pi i}{N_n}\right), \\ y_{s_i} &= y_0 + r \cdot \sin\left(\frac{2\pi i}{N_n}\right), \end{aligned} \qquad (14)$$

with $N_n$ representing the number of samples. However, some test points obtained from expression (14) may not be taken into account despite actually belonging to the circle boundary as a consequence of the uniform sampling based on $N_n$.